\documentclass[11pt,a4paper]{article}

\usepackage[hyperref]{acl2021}
\usepackage{times}
\usepackage{latexsym}
\usepackage{todonotes}
\usepackage{amsmath}
\usepackage{amsfonts}
\usepackage[noend]{algpseudocode}
\usepackage{algorithm}
\usepackage{tabularx}
\usepackage{booktabs}
\usepackage{hyperref}
\usepackage[frozencache=true,cachedir=minted-cache]{minted} 
\usepackage{listings}

\usepackage{microtype}
\newcommand{\E}{\mathbb{E}}

\newcommand{\KL}{D_{\mathrm{KL}}}

\newcommand{\pit}{{\pi_\theta}}

\aclfinalcopy % Uncomment this line for the final submission
\def\aclpaperid{13} %  Enter the acl Paper ID here

\title{Energy-Based Models for Code Generation \\under Compilability Constraints}

\author{Tomasz Korbak,$^{1,}$\thanks{\hspace{0.2cm}Work done during a research internship at Naver Labs Europe.} \ Hady Elsahar,$^2$ Marc Dymetman,$^2$ Germán Kruszewski$^2$\\
  \texttt{t.korbak@sussex.ac.uk}\\ \texttt{\{hady.elsahar,marc.dymetman,german.kruszewski\}@naverlabs.com}\\
  $^1$University of Sussex, United Kingdom \\
  $^2$Naver Labs Europe, France\\
}

\date{}

\begin{document}
\maketitle
\begin{abstract}
Neural language models can be successfully trained on source code, leading to applications such as code completion. 
However, their versatile autoregressive self-supervision objective overlooks important global \textit{sequence-level} features that are present in the data such as syntactic correctness or compilability. 
In this work, we pose the problem of learning to generate compilable code as constraint satisfaction. 
We define an Energy-Based Model (EBM) representing a pre-trained generative model with an imposed constraint of generating only compilable sequences. 
We then use the KL-Adaptive Distributional Policy Gradient algorithm~\cite{khalifa_2021} to train a generative model approximating the EBM. We conduct experiments showing that our proposed approach is able to improve compilability rates without sacrificing diversity and complexity of the generated samples. 
\end{abstract}

\section{Introduction}

% code completion why is useful 
Code completion is an essential feature of any modern Integrated Development Environment (IDEs). It supports developers with recommendations about the next token to write given a context,  speeding up software development and reducing the number of mistakes. 
A large body of work has relied on statistical language modeling, treating programming languages as natural languages using probabilistic grammars ~\cite{Raychev2014,Bielik2016}, and more recently relying on neural language models~\cite{ncc_liu2016,Svyatkovskiy,ncc_Svyatkovskiy2020,ncc_Arkesteijn2020,ncc_Ciniselli2021}.\footnote{See~\citet{survey_AllamanisBDS18} for a survey.}
%% neural models are prominent 
In particular, neural autoregressive language models have been  favoured due to their scalability and generic training procedure that can exploit large codebases (e.g. open source code repositories available on GitHub) through self-supervised training. 
%% neural models oversee priors that might be important 

Despite these desirable traits, neural language models, trained in the standard way, are known to suffer from myopia and to overlook global sequence-level features that are present in the data and which might be crucial for the quality of generated sequences~\cite{A-parshakova-etal-2019-global}.
This leads to repetitions, hallucinations and failing to capture long-distance consistency requirements. In a code generation context, this is demonstrated in compilation errors that are a common failure mode in such tasks as translation between programming languages~\citep{roziere2020unsupervised}.
This problem has inspired a large body of work on different fronts on injecting sequence-level priors by either directly optimizing sequence-level features~\cite{seq_lvl_train_RanzatoCAZ15} or through fusion with grammars and automata~\cite{chunyang}. These techniques aim to balance between the desirable traits and fast inference of neural autoregressive models trained in the standard way and the satisfaction of global sequence-level features.

% Sequence level optimization
% RL methods maximize reward on the expense of diversity 
In this work, we formulate compilable code generation as a constraint satisfaction problem. We show that this formulation leads to a unique distribution represented by an Energy-Based Model (EBM). This unique distribution by definition fully satisfies the compilability constraints while having a minimal KL divergence from the original autoregressive generative model trained through cross entropy.
% KL DPG
We then train an auto-regressive generative model to approximate the underlying distribution of this EBM using the KL-Adaptive Distributional Policy Gradient algorithm~\cite{khalifa_2021}. 

In our experiments, we show that our approach significantly improves compilability rates without sacrificing diversity or complexity of the generated examples. 
This alleviates the drawbacks of reinforcement learning fine-tuning techniques that maximize compilability but deviate significantly from the original generative model, which leads to severe loss in diversity and complexity of the generated samples.
Finally, we complement our experiments with a qualitative analysis of the effect of several fine-tuning approaches on the distribution of compilation errors.
\\

%
% Overall, the main contributions of this work can be summarized as follows:
% \begin{itemize}
%     \item We formalize compilable code generation as a constraints satisfaction problem withing the recently proposed GDC framework~\cite {}.
%     \item We empirically show that GDC can increase compilation rate while maintaining quality, diversity and low perplexity.
%     \item We conduct quantitative and qualitative error analysis, examining how different fine-tuning approaches affect different categories of compilation errors 
% \end{itemize}

\section{Related Work}

\paragraph{Imposing compilability constraints on generative models} There is a body of work focusing on unconditional code generation or code completion: generating a piece of source code given a preceding piece of source code \cite{Nguyen2013,Raychev2014,Karpathy2015VisualizingAU,Bielik2016}. That work, however, focuses on perplexity and similarity with respect to ground truth completions (in terms of exact-match accuracy, Levensthein distance and ROUGE scores) \cite{Svyatkovskiy,codexglue}, usually failing to measure and control for compilability of generated sequences or semantic and syntactic constraints in general.\footnote{One exception is the work of \citet{Maddison2014}, who augment neural probabilistic context free grammars with semantic constraints and use them for unconditional generation.} On the other hand, semantic and syntactic constraints are frequently considered in language-to-code translation or program synthesis. For instance, \citet{zhong2017seq2sql}, who used policy gradients to train a model for translating natural language questions to corresponding SQL queries and -- in addition for rewarding for query execution results -- added a penalty for syntactically invalid queries. Taking that one step further, \citet{Kulal2019} use compilation errors (with their precise location) to guide search over the space of possible programs.

\paragraph{Optimizing sequence-level rewards for text generation} 
Most previous attempts at steering autoregressive model to conform to global constraints defined over entire sequence have employed reinforcement learning (RL). 
This includes using Reinforce~\citep{Williams92} for machine translation \cite{seq_lvl_train_RanzatoCAZ15} or actor critic~\cite{conda_actor} for abstractive summarization~\citep{PaulusXS18}, caption generation~\cite{RL_Img2txt_LiuZYG016}, dialogue~\cite{RL_dialogue_LiMRJGG16}, and video captioning~\citep{PasunuruB17}. Some approaches (for instance, in machine translation and summarization \citep{seq_lvl_train_RanzatoCAZ15, BahdanauBXGLPCB17}) directly optimize performance metrics such as BLEU and ROUGE at training time. 
Others use heuristic rewards (for instance \citet{RL_dialogue_LiMRJGG16} for dialogue generation and \citet{RL_TambwekarDMMHR19} for story generation) in order to obtain certain a priori desirable features of generated sequences that then incentivize good performance on target metrics. 
A weakness of using RL in fine-tuning generative models is the problem of catastrophic forgetting: maximizing global, sequence-level rewards leads to very large deviations from the original autoregressive model trained through cross-entropy. This often results in significant reductions in fluency and diversity of generated samples. 
The catastrophic forgetting problem is sometimes addressed by imposing a penalty term to the rewards, such as the KL divergence between the trained policy and the auto-regressive model. This approach, termed ``conservative fine-tuning", was applied to generating melodies with music theory rewards and organic molecules with synthesizability rewards by \citet{Jaques-2017} as well fine-tuning language models for controllable language generation by \citet{Ziegler19}. This solution doesn't have an explicit notion of the optimal policy and often has hard time balancing between the reward term and the KL penalty term, leading to instability in training ~\cite{khalifa_2021}. Unlike this approach, our formulation defines the optimal distribution that satisfies both requirements.

\paragraph{Energy-based models for text}
Energy-based models (EBMs)~\citep{Hinton02,lecun_tutorial_2006,RanzatoBCL07} are a family of probabilistic graphical models in which learning and inference are done by associating an unnormalized probability with each configuration of observed and latent variables. Early examples of EBMs applied to natural language processing include sequence labeling problems (e.g. tagging) exploiting global properties of a  sequence~\citep{andor_globally_2016,Belanger:2016:SPE:3045390.3045495}. A recent surge of interest in EBMs \citep{mordatch_ebms} has not left text generation unaffected (see \cite{Bakhtin2020EnergyBasedMF} for a survey). \citet{Tu2020ENGINEEI} proposed an energy-based inference networks for non-autoregressive machine translation. \citet{A-parshakova-etal-2019-global} and \citet{Deng_EBM_20} augment a autoregressive language models with an additional global factor to obtain a lower perplexity on the training data. \citet{khalifa_2021} develop a novel approach to distributional controllable text generation by constructing an EBM satisfying desired statistical constraints imposed on the set of generated sequences (such as topic or gender statistics over the sequences) and then train an autoregressive policy to approximate it, which can be sampled from efficiently. We build on \citeauthor{khalifa_2021}'s approach by applying it to a novel domain outside natural language and defining a new kind of constraint: compilability.

\section{Method}

Following \citet{khalifa_2021}, we formulate compilable code generation as a constraint satisfaction problem over a space of generative models. 
There are two constraints that a target generative model $p$ must satisfy. First, $p$ must have minimal divergence -in the distribution space- from an original generative model $a$ pre-trained using a standard autoregressive language modeling objective. Second, it must generate only sequences that satisfy a certain sequence level constraint $b$. In our case, $b(x) = 1$ iff $x$ is a syntactically correct Python program and $b(x) = 0$ otherwise.
There two constraints can be represented as a product-of-experts \citep{Hinton02} energy-based model
\begin{equation}
    P(x) = a(x)b(x).
\end{equation}
$p(x)$ can be obtained from $P(x)$ by dividing it by a normalization constant $Z$:
\begin{equation}
    p(x) = \frac{1}{Z} P(x),
\end{equation}
where
\begin{equation}
    Z \doteq \sum_x P(x).
\end{equation}
This EBM $P$ is unique, it represents a distribution $p$ that optimally reconciles the two constraints. It is a special case of the generalized maximum entropy formulation presented in~\citep{csizarShields2004} for applying constraints over distributions.

However, one problem still remains: it is not straightforward how to draw samples $x \sim p(x)$ or even evaluating probability $p(x)$ from this optimal unique distribution.
A simple method for drawing samples from the $p$ distribution could be sampling sequences from $a$ and filtering on $b(x)$. While this method sounds simple, there's no direct way of using it for interactive code completion as sampling full sequences till the end is necessary to filter through the sequence-level filter $b(x)$. 
Therefore our objective here is to obtain another autoregressive policy $\pit$ to directly approximate $p$. 

To attain this, \citet{khalifa_2021} (following \citet{opt-rl-arxiv-2019}) developed a training procedure called KL-Adaptive Distributional Policy Gradients (KL-DPG) to train $\pit$ to minimize the KL divergence between $p$ and $\pit$. The gradient of this KL turns out to be tractable:
\begin{align}
% \begin{split}
\label{eq:gradKL}
\nabla_\theta \KL(p, \pit) &= \nabla_\theta \E_{x \sim p} \log \frac{p(x)}{\pit(x)} \\
&= - \nabla_\theta \E_{x \sim p} \log \pit(x) \\
&= - \E_{x \sim p} \nabla_\theta \log \pit(x) \\
&= - \frac{1}{Z}\sum_x P(x) \nabla_\theta \log \pit(x)
% \end{split}    
\end{align}
Let us now absorb the constant $-1/Z$ into a learning rate $\alpha^{(\theta)}$ and estimate the expectation over $p(x)$ using importance sampling \cite{owen_chapter_importance_sampling_2013} from yet another generative model $q$:
\begin{equation}
    \label{eq:dpg-off}
    \nabla_\theta \KL(p, \pit) \propto \E_{x \sim q} \frac{P(x)}{q(x)} \nabla_\theta \log \pit(x).
\end{equation}
During training, both $\pit$ and $q$ are initialized as $a$. Then, $q$ is periodically updated to $\pit$ if $\pit$ surpasses $q$ in being closer to $p$ (in terms of KL). For a pseudo-code of the whole KL-DPG training procedure, see Algorithm \ref{al:KL-adaptive-DPG}.

The gradient in \eqref{eq:dpg-off} is similar to an estimate obtained using policy gradients methods in standard reinforcement learning \cite{sutton_policy_gradients} with $P(x)/q(x)$ playing the role of a pseudoreward. This similarity, however, is superficial. Our objective is approximating a target generative model $p$ by minimizing $\KL(p, \pit)$ rather than maximizing expected reward $b(x)$ or $P(x)$ or $P(x)/q(x)$. As we show in Section \ref{sec:results}, these objectives produce vastly different policies which diverge from $p$ and catastrophically forget what the pretrained model $a$ knew about its training domain. Furthermore, since $q$ will always be close to $\pit$, our pseudoreward $P(x)/q(x)$ effectively depends on policy parameters $\theta$.

\begin{algorithm}[H]
\caption{KL-DPG \label{al:KL-adaptive-DPG}}
\begin{algorithmic}[1]
\Require EBM $P$, initial generative model $a$
\State $\pi_\theta \gets a$
\State $q \gets a$
\For{each iteration}
\For{each episode}
    \State sample $x$ from $q(x)$
    \State $\theta \gets \theta + \alpha^{(\theta)} \frac{P(x)}{q(x)} \ \nabla_\theta \log \pi_\theta(x)$ 
\EndFor
\If{ $\KL(p||\pi_\theta) <  \KL(p||q)$} 
    \State $q \gets \pi_\theta$
\EndIf
\EndFor
\Ensure $\pi_\theta$
\end{algorithmic}
\end{algorithm}

\section{Experiments}
\label{sec:experiments}
\subsection{Setup}
\paragraph{Dataset:}
To prepare the training dataset,  we started from the Python150 dataset, which consists of 150k Python source code files obtained from GitHub \cite{Raychev2016}. Then, using the code from \citet{roziere2020unsupervised}, we extracted 713k Python functions (both methods and standalone functions) from it (250 MB of raw text data). The additional filtering criteria were compilability (according to $b(x)$) and being less than 128 BPE tokens long. The dataset was then split into a training subset $\mathcal{D}_\text{train}$ and test subset $\mathcal{D}_\text{test}$.
\paragraph{Initial generative model $a$:}
We implemented $a$ using the GPT-2 \cite{radford2019language} architecture with 117m parameters (\texttt{gpt2-small}) and kept all the original hyperparameters (see Table \ref{table:hyperparams-gpt} in the Appendix). We trained a byte-level BPE tokenizer \cite{sennrich-etal-2016-neural} with special BOS and EOS tokens to obtain a vocabulary of 50k tokens. The model was trained for one epoch.

\paragraph{Compilability Scorer $b$:}
To check for compilability, we call the \texttt{compile\_command} function from \texttt{codeop} module of Python Standard Library\footnote{\url{https://docs.python.org/3/library/codeop.html}} with a sequence $x$ as argument and check if it returns a \texttt{code} object. We apply no postprocessing other than removing BOS and EOS tokens. 
\texttt{codeop.compile\_command} is the implementation that Python interactive interpreters use in read-eval-print loop (REPL) to determine whether a string is a valid Python code. The method tries to compile a string of Python code and raise and exception if there is a problem with the Python code, in particular 
a \texttt{SyntaxError} for invalid Python syntax and \texttt{ValueError} or \texttt{OverflowError} if there is an invalid literal.  

This notion of compilability is concerned only with syntactic correctness and does not execute the body of a function. However, we found the initial compilability rate $\E_{x \sim a}b(x)$ of functions $x$ sampled from $a(x)$ to be only 0.56, which leaves a large margin for improvement.\footnote{Note that initial compilability rate will be equal to our $Z$ because $\E_{x \sim a}b(x) = \sum_x a(x)b(x) = \sum_x P(x) = Z$.}

\paragraph{KL-DPG training}

$\pit$ and $q$ share their architecture with $a$ but have separate weights which are only initially identical to $a$'s. Throughout the training, $\pit$ will be updated to approximate $p$. See Table \ref{table:hyperparams-gdc} in the Appendix for a complete list of hyperparameters used for training $\pit$ and $q$ using KL-DPG.

\subsection{Baselines}

We compare our method to a common approach of using standard reinforcement learning to fine-tune a generative model to conform to desired constraints. We use the Reinforce algorithm \citep{Williams92Reinforce} which instead of minimizing divergence from the target distribution $p$ tries to maximize expected reward $\E_{\pit} R(x)$. We consider two kinds of reward $R(x)$:
\begin{itemize}
    \itemsep0em 
    \item $R(x) = b(x)$, where the generative model is simply rewarded for generating sequences that compile;
    \item $R(x) = P(x)$, where the generative model is simply rewarded proportionally to the score our EBM assigns to $x$. Intuitively, this objective gives reward for both compilability and respecting the  original generative model $a$.
\end{itemize}

\subsection{Evaluation Metrics}
We evaluate KL-DPG and two baselines in terms of the following metrics:
\begin{enumerate}
    \itemsep0em 
    \item $\E_{x \sim \pit} b(x)$, compilability rate of sequences sampled from $\pit(x)$,
    \item $\KL(p, \pit)$, the forward KL divergence from the optimal distribution $p$,
    \item $\KL(\pit, a)$, the reverse KL divergence from the original pretrained generative model,
    \item Distinct-1 score, a measure of text diversity in terms of the frequency of token repetitions in a sample $x$, proposed in the context of NLP by \citep{li-etal-2016-diversity},
    \item Self-BLEU-5, a measure of text diversity \emph{across} samples, proposed in the context of NLP by \citep{texygen-ZhuLZGZWY18},
    \item Perplexity measured on $\mathcal{D}_\text{test}$, a held-out subset of the data used for training $a$, calculated as 
    $$
    \exp \Big [ - \frac{1}{N}  \sum_{x \in \mathcal{D}_\text{test}} \log \pit(x) \Big],
    $$
    where $N$ is the overall number of tokens in $\mathcal{D}_\text{test}$.
    \item Sequence length, the average number of characters in generated sequence $x$ after detokenization,
    \item AST node count, the average number of nodes in an abstract syntax tree (AST) of sequences that compile. Samples are parsed to their corresponding ASTs using the \texttt{ast} module from Python Standard Library.\footnote{\url{https://docs.python.org/3/library/ast.html}} Intuitively, this metric should indicate the logical (as opposed to surface) complexity of generated programs,
    \item PEP8 error frequency, the average  number of violations of PEP8, the style guide for Python,\footnote{\url{https://www.python.org/dev/peps/pep-0008/}} measured using pycodestyle,\footnote{\url{https://github.com/PyCQA/pycodestyle}} an off-the-shelf linter (static code analysis tool).  We report the average number of errors per character to avoid confounding by sequence length.
\end{enumerate}

While high compilability rate is the target, the remaining metrics control for various aspects of fluency, quality and diversity of generated samples. Most but not all of these aspects reduce to the constraint of staying close to $a$; for instance, it is possible for $\pit$ to actually outperform $a$ in matching the statistics of $a$'s own training distribution $p^*(x)$.
%%%%%%%%%%%%%%%%%%%%%%%%%%%%%%%%%%
%%%%%%%%%%%% RESULTS %%%%%%%%%%%%%
%%%%%%%%%%%%%%%%%%%%%%%%%%%%%%%%%%
\begin{figure}[ht]  
    \centering
    \includegraphics[width=\linewidth]{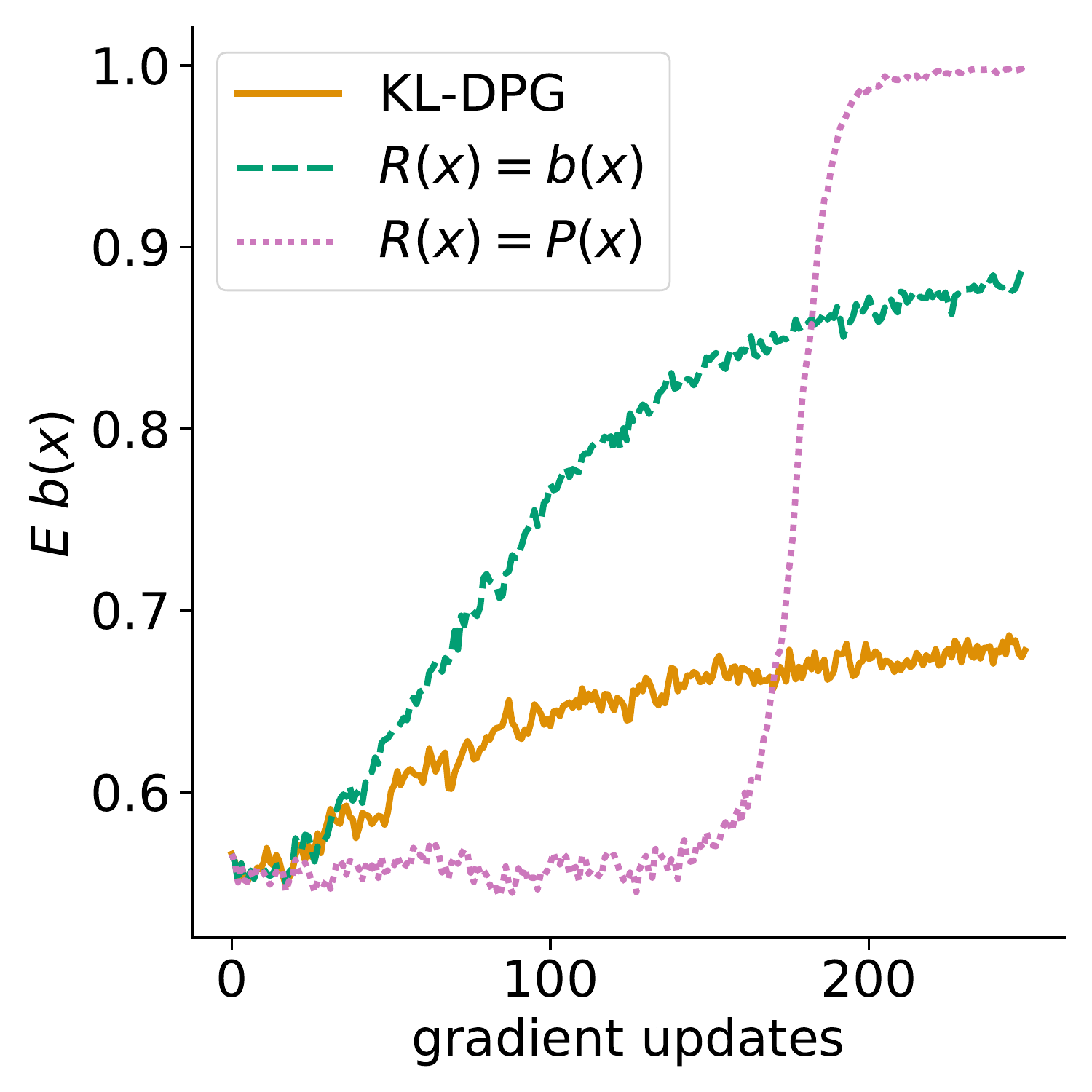} 
    \caption{\small{Compilability rate $\E_{{x \sim \pit} b(x)}$ ($\uparrow$ better) of samples from policies obtained from KL-DPG, and two baselines: Reinforce with reward $R(x) = b(x)$ and with reward $R(x) = P(x)$.}}
    \label{fig:cr}
\end{figure}

\begin{figure*}[ht]  
    \centering
    \includegraphics[width=\linewidth]{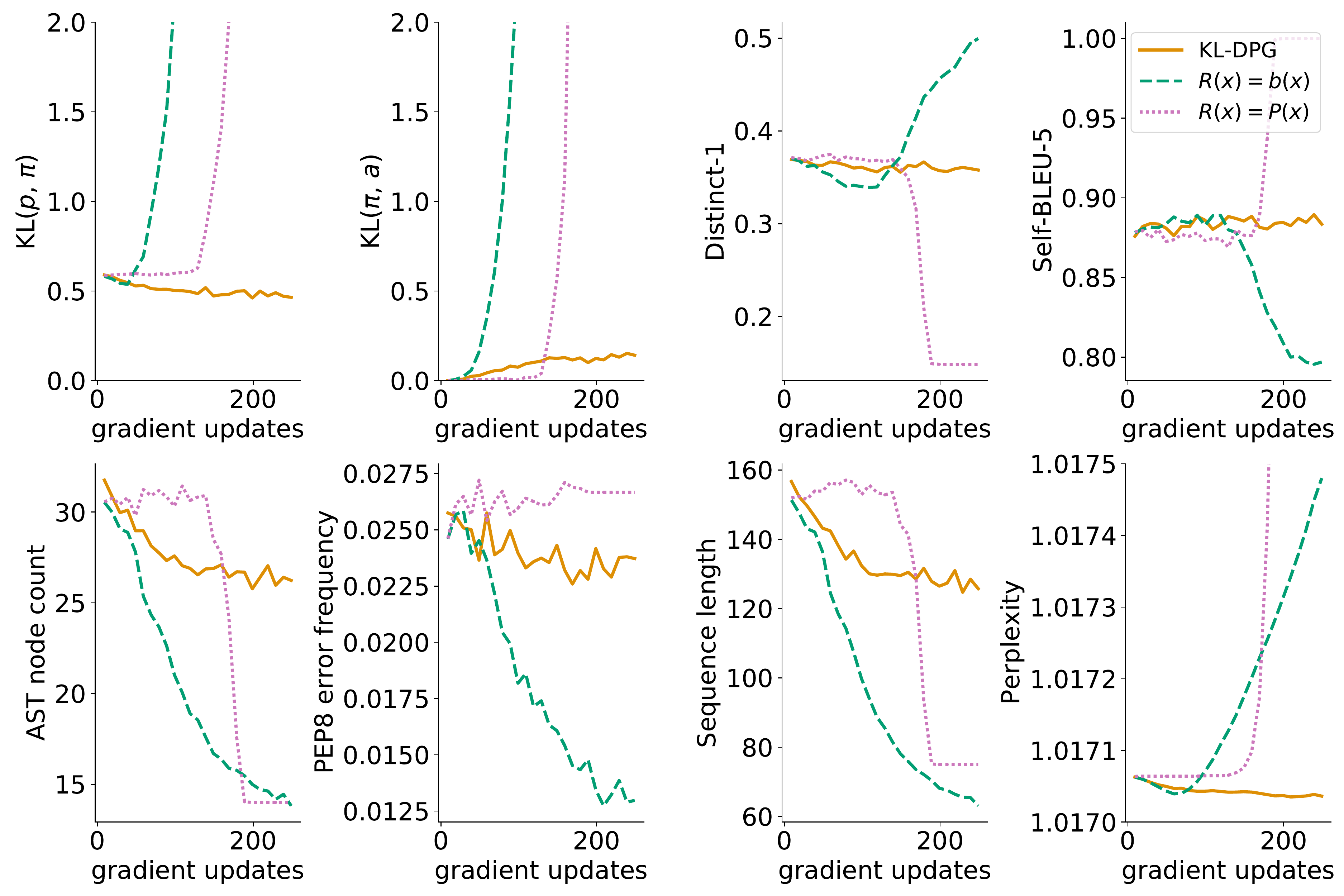} 
    \caption{\small{Evaluation metrics $\text{KL}(p|\pi_{\theta})$ ($\downarrow$ better), $\text{KL}(\pi_{\theta}|a)$ ($\downarrow$ better), Self-BLEU-5 ($\downarrow$ better), Distinct-1 ($\uparrow$ better), AST node count ($\uparrow$ better), PEP8 error count ($\downarrow$ better), sequence length ($\uparrow$ better), and perplexity ($\downarrow$ better) for policies obtained from KL-DPG, and two baselines: Reinforce with reward $R(x) = b(x)$ and with reward $R(x) = P(x)$.}}
    \label{fig:main}
\end{figure*}
%%%%%%%%%%%%%%%%%%%%%%%%%%%%%%%%%%
\begin{figure*}[ht]  
    \centering
    \includegraphics[width=\linewidth]{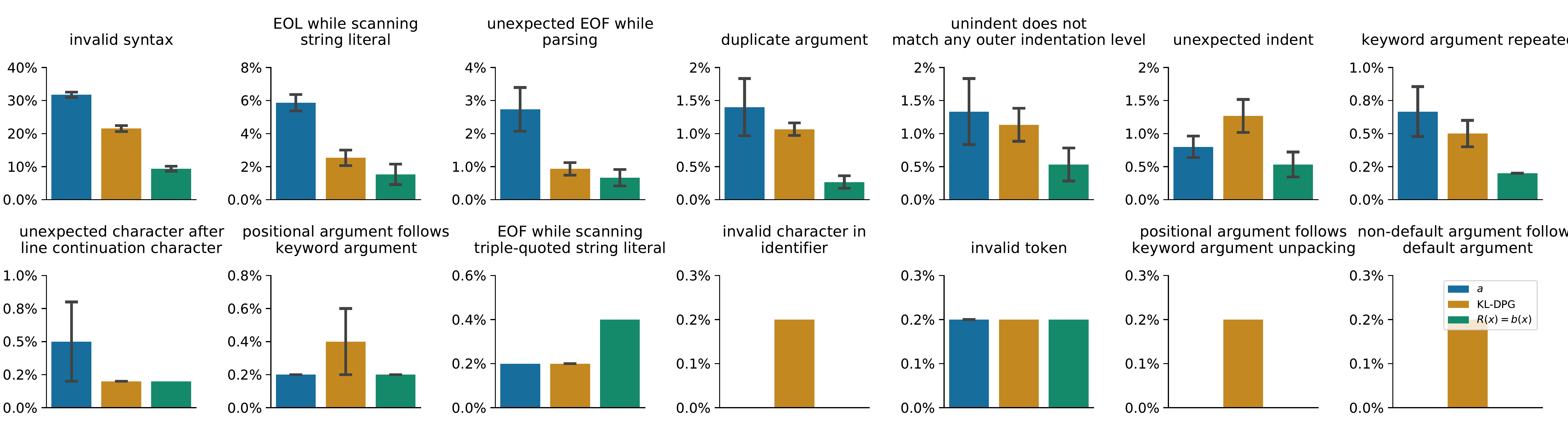} 
    \caption{\small{The frequency (measured as the percentage of samples from $\pit(x)$ causing a given error) of each kind compilation error for the original generative model $a$ and policies fine-tuned using KL-DPG and Reinforce with $R(x) = b(x)$. The policy fine-tuned using Reinforce with $R(x) = P(x)$ was excluded because the single sequence it produces causes no compilation errors. Percentages were computed using 500 samples while confidence intervals were based on 3 repeats of the sampling procedure.}}
    \label{fig:errors}
\end{figure*}
%%%%%%%%%%%%%%%%%%%%%%%%%%%%%%%%%
\section{Results}
\label{sec:results}
We present the evolution of nine evaluation metrics as a function of gradient updates on Figures~\ref{fig:cr}~and~\ref{fig:main}.

Reinforce with $R(x) = b(x)$ quickly improves compilability by a large margin but this improvement is mirrored by an equally large divergence from $p$ and $a$. This divergence translates into generating sequences much shorter (in terms of the number of characters) and logically simpler (in terms of the number of nodes in its AST) than an average sequence sampled from $a$. This heavily decreased sequence length (most of the generated functions are one-liners) seems to artificially increase diversity metrics (Self-BLEU-5 and Distinct-1).

Reinforce with $R(x) = P(x)$ doesn't improve compilability rate until an inflection point after which it quickly reaches perfect compilability at a price of heavily diverging from both $a$ and (perhaps counterintuitively) $p$. The reason behind that, however, is that the policy heavily peaks around a single sequence that is compilable. To understand what causes this behavior, first note that the objective for Reinforce with $R(x) = P(x)$ is to maximize $\mathbb{E}_{x \sim \pit} [a(x)b(x)]$. Because $R(x) = 0$ for uncompilable sequences, compilation rate will improve. But for compilable sequences, the effective reward is $R(x) = a(x)$ meaning that $\pit$ is rewarded most for generating the most probable sequences (according to $a(x)$), making them even more probable. Eventually, $\mathbb{E}_{x \sim \pit} a(x)$ is maximized by a policy peaking on a single sample $x$ that was the most probable one according to $a(x)$. This failure mode is reflected in diversity metrics and perplexity. The sequence the policy peaks on is also shorter and less complex than an average sequence sampled from $a$.

KL-DPG is the only method that consistently improves compilability rate while decreasing divergence from $p$, maintaining the diversity of $a$ and only slightly decreasing sequence length and the number of nodes in ASTs. Moreover, as a by-product of improving compilability, KL-DPG is also able to slightly decrease the perplexity and the frequency of PEP8 violations per character. We conjecture the decrease in perplexity is because compilability provides a training signal enabling $\pit$ to fit the $a$'s training distribution $p^*(x)$ better than $a$ was able to.\footnote{This mirrors the results obtained by \citet{A-parshakova-etal-2019-global}, who also defined an EBM augmenting an autoregressive model with prior knowledge about features of the training set and observed a decrease in perplexity compared to pure autoregressive training.} The decrease in the frequency of PEP8 violations might be due to the fact that compilability is correlated with PEP8 compliance.

\subsection{Qualitative evaluation}

\begin{figure}[ht]  
    \centering
    \includegraphics[width=\linewidth]{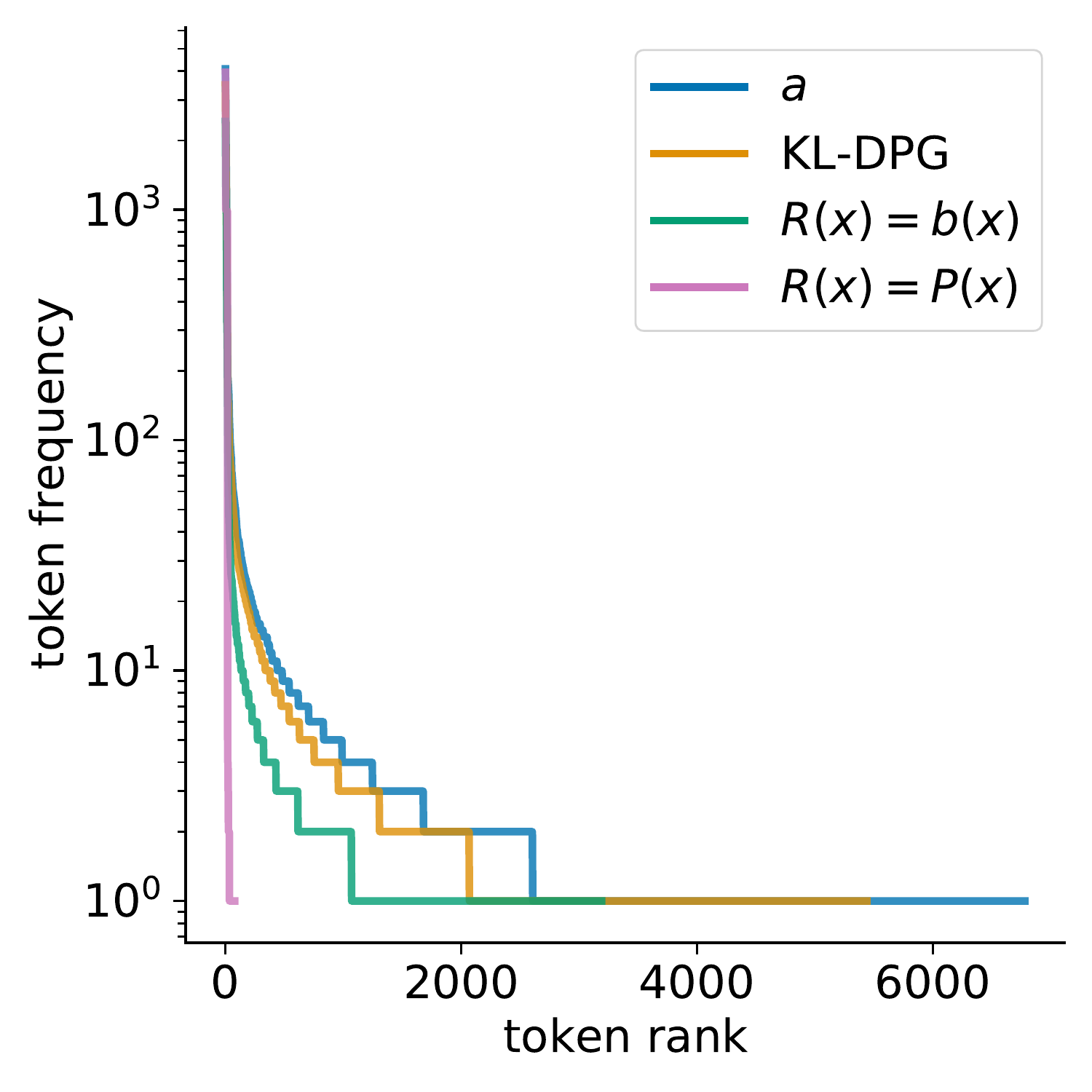} 
    \caption{\small{Token frequency against token rank computed for tokens found in samples from from KL-DPG, and two baselines. Longer tails imply more diverse samples.}}
    \label{fig:zipf}
\end{figure}

To further analyze effects of different fine-tuning approaches on sample diversity, we measured the frequency of BPE tokens in generated samples. For each of four analyzed generative models, we sampled 1000 sequences using pure ancestral sampling. We then computed the frequency for each BPE token (the number of times it occurs) and its rank (its index in a sorted list of tokens). We plotted these results on Figure \ref{fig:zipf}. This qualitative evaluation paints a similar picture: fine-tuning using Reinforce incurs a large (with $R(x) = b(x)$) or extreme (with $R(x) = P(x)$) decrease in token diversity. In contrast, KL-DPG is able to maintain a relatively long tail of token frequencies, not departing too far from $a$.

Moreover, in order to gain better understanding of how different fine-tuning methods affect generative models we measured the frequency of different categories of compilation errors for samples from $a$ and from fine-tuned policies. This analysis is presented on Figure \ref{fig:errors}. We categorized errors using error messages produced by Python interpreter trying to compile an uncompilable sequence. \texttt{invalid syntax} is the most common failure mode (30\% of all sequences sampled from $a$), with a long tail of other error categories. We can see that both KL-DPG and Reinforce with $R(x) = b(x)$ consistently decrease error frequency across almost all the categories.

Finally, in the Appendix we present randomly generated samples from each discussed policy. Tables \ref{fab:a_samples}-\ref{fab:reinforceP_samples} contain samples obtained through unconditional generation. In addition to that, to illustrate the applicability of obtained policies for code completion, in Tables \ref{tab:def close_samples}-\ref{tab:def generate_samples_with_prompt_samples} we present samples obtained through conditional generation, i.e. $x \sim \pit(x|c)$, where the context $c$ is a function name. In either case, samples were obtained using pure ancestral sampling.

\section{Discussion}
In the paper, we presented a new energy-based model formulation for the problem of imposing the constraint of compilability on an autoregressive generative model for source code. 
In contrast with standard reinforcement learning approaches, the solution we propose -- KL-DPG -- is able to improve compilability rate without sacrificing diversity and complexity of generated samples.

One obvious application of the presented approach is improving the accuracy of code completion, i.e. tools assisting in programming by predicting the next tokens based on context \citep{Svyatkovskiy}. The fact that fine-tuning using KL-DPG has a beneficial effect on perplexity and PEP8 error frequency suggests that it can provide a training signal complementary to that in a language modeling objective. The benefits of this auxilary training signal would arguably diminish with increased training time and datatset size, but that still leaves room for significant improvement in low-resource domains.

A limitation of the current KL-DPG approach is that it is restricted to unconditional generation. This is because for a conditional EBM $P(x,c)$ the proportionality constant $-1/Z$ from \eqref{eq:gradKL} would depend on a context $c$. Nevertheless, one can imagine using a policy $\pit$ fine-tuned using KL-DPG as initialization of a decoder for conditional generation, e.g. transpilation (translation between programming languages) or program synthesis (translation from a natural language to a programming language).
\bibliographystyle{acl_natbib}
\bibliography{anthology,additional_references}

%\iffalse % comment appendix provisionally
\clearpage
\appendix

\section{Hyperparameters and implementation details}
\label{appendix:Hyperparameters}

We implemented all models using PyTorch \citep{pytorch} and HuggingFace \citep{huggingface}. Training the initial generative model took 10 days on 3 Nvidia Tesla T4 GPUs. For a detailed list of  hyperparameter values, see Table \ref{table:hyperparams-gpt}.

% ; for description of hyperparameters specific to Ziegler see \citep{Ziegler19}.

\begin{table}[H]
    \footnotesize
    \centering
    \begin{tabular}{l|l}
    \toprule
    \textbf{Hyperparameter} & \textbf{Value}  \\
    \toprule
    base LM & \texttt{gpt2-small} \\
    number of params & 117m \\
    number of layers & 12 \\
    number of heads & 12 \\
    vocabulary size & 50257 \\
    sequence length & 128 \\
    hidden state size & 768 \\
    activation function & gelu \\
    optimizer & Adam \citep{kingma2014adam}\\
    initial learning rate & $5 \times 10^{-5}$ \\
    learning rate scheduler & linear \\
    batch size & 24 \\
    total gradient updates & 20069 \\
    dropout rate & 0.1 \\
    \bottomrule
    \end{tabular}
    \caption{Hyperparameters used for training the initial generative model $a$}
    \label{table:hyperparams-gpt}
\end{table}

The implementation of KL-DPG was based on code published by \citet{khalifa_2021}.\footnote{\url{https://github.com/naver/gdc}} Each fine-tuning run took approximately 5 days on 2 Nvidia V100 GPUs. For a detailed list of hyperparameter values, see Table \ref{table:hyperparams-gdc}.

\begin{table}[H]
    \footnotesize
    \centering
    \begin{tabular}{l|l}
    \toprule
    \textbf{Hyperparameter} & \textbf{Value}  \\
    \toprule
    % \multicolumn{2}{l}{\textbf{Common hyperparameters}} \\
    optimizer & Adam \citep{kingma2014adam}\\
    learning rate $\alpha^{(\theta)}$ & $1.41 \times 10^{-6}$ \\
    learning rate scheduler & linear \\
    batch size & 2048 \\
    warmup gradient updates & 100 \\
    total gradient updates & 250 \\
    sequence length & 128 \\
    dropout rate & 0.1 \\

    % \multicolumn{2}{l}{\textbf{Hyperparameters specific to Ziegler}} \\
    % $\gamma$ & 1 \\
    % $\lambda$ & 0.95 \\
    % clip range & 0.2 \\
    % target KL & 6.0 \\
    % initial KL coefficient & 0.2 \\
    % horizon & $10^4$ \\
    \bottomrule
    \end{tabular}
    \caption{Hyperparameters used for training $\pit$ using KL-DPG and Reinforce}
    \label{table:hyperparams-gdc}
\end{table}

\begin{table*}[t]
\tiny
\begin{tabular}{l|l}
\toprule
\textbf{$b(x)$} & \textbf{Program} \\ 
\midrule
0 & 

% \begin{minipage}{.9\textwidth}
% \begin{minted}[breaklines]{python}
% def SpecificCodeCPP(self):
%     s = "\nGEOnerry API:\n  tickets while path %s" % self.to_read(s))
%     s += "\n. post\n"
%     resp = 201 + '\n\n  L, * ok(resp)
%     return resp

% \end{minted}
% \end{minipage}
% \\
% \\
% \hline
% \\
% 1 & 

% \begin{minipage}{.9\textwidth}
% \begin{minted}[breaklines]{python}
% def __init__(self,value,key,parent = None):
%     self.value = value
%     self.item = item
%     self.parent = parent

% \end{minted}
% \end{minipage}
% \\
% \\
% \hline
% \\
% 0 & 

\begin{minipage}{.9\textwidth}
\begin{minted}[breaklines]{python}
def test_3_invalid(self):
    serializer = serializer.validated_manager['quarterly_ cred']
    serializer.user = 'token'
    self.verify_token(epsg = serializer.DBModes,[serializer.user])

\end{minted}
\end{minipage}
\\
\\
\hline
\\
0 & 

\begin{minipage}{.9\textwidth}
\begin{minted}[breaklines]{python}
def delete(self,username,password = None):
    if username:
        if username.startswith("oil",None)or username.startswith('"",True):
            raise HttpRequest()
    db.model.delete.assert_called_with(username,'password')

\end{minted}
\end{minipage}
\\
\\
\hline
\\
1 & 

\begin{minipage}{.9\textwidth}
\begin{minted}[breaklines]{python}
def mode(self):
    self._mode = 'modeM_GB'
    return self

\end{minted}
\end{minipage}
\\
\\
\hline
\\
0 & 

\begin{minipage}{.9\textwidth}
\begin{minted}[breaklines]{python}
def _update_update_tbl(self,new_worksheet):
    self._merge_tbl(new_worksheet,old_worksheet)
    self._create_where('x1')
    self._update_tbl('x1',{ }).extend([str(new_fh.getvalue()))
    self._clear_sql()
    self.clear_lstrip()
    self.pop.set('x1')[int(col)for param in['x1','y1']]
    self.flush.update()

\end{minted}
\end{minipage}
\\
\\
\hline
\\
1 & 

\begin{minipage}{.9\textwidth}
\begin{minted}[breaklines]{python}
def _callResourceCost(self,server):
    response = urllib.Request('GET','//api//log//%s//detected//' % server.id)
    body = urllib. urllib2.urlencode(body)
    response.headers['X-Basic-Control-Authorization']= self.oauth_client.Client.CertResponse(response.body)
    return response

\end{minted}
\end{minipage}
\\
\\
\hline
\\
0 & 

\begin{minipage}{.9\textwidth}
\begin{minted}[breaklines]{python}
def _pre_save(self,data):
    self.calculate_updates([item.resolve(data['output')]= yield data['output'].find('top',['mybounce','geodeIB'])))

\end{minted}
\end{minipage}
\\
\\
\hline
\\
1 & 

\begin{minipage}{.9\textwidth}
\begin{minted}[breaklines]{python}
def read(self):
    self.offset -= 1
    start = O8(self)
    while time.time()- start:
        return self.get_index(start)

\end{minted}
\end{minipage}
\\
\\
\hline
\\
0 & 

\begin{minipage}{.9\textwidth}
\begin{minted}[breaklines]{python}
def Pub(self):
    r = PCHAP()
    r['where']= struct.unpack('!T',self.digest))
    response = MKchronosOPS('R')
    self.sendMessage(response)
    return self.Response(response)

\end{minted}
\end{minipage}
\\
\\
\hline
\\
1 & 

\begin{minipage}{.9\textwidth}
\begin{minted}[breaklines]{python}
def __init__(self,current_node):
    self.current_node = current_loadbalancer
    self.assign_current_node = None
    self.parenting = None
    if self.menu:
        self.getNodeSelector(Index(RemovelineToRow,self.parent.position),0,2.0,5.0)
    self.show_parent()

\end{minted}
\end{minipage}
\\
\\
\hline
\\
1 & 

\begin{minipage}{.9\textwidth}
\begin{minted}[breaklines]{python}
def get_response_data(self):
    return { 'from_blob_client':self.to_blob_key,'as_blob_secret':self.to_project_secret.to_secret(),'json':self.to_storage }

\end{minted}
\end{minipage}
\\
\\
\hline
\\
0 & 

\begin{minipage}{.9\textwidth}
\begin{minted}[breaklines]{python}
def put(self,key,expire = True):
    if not invert:
        dict = { }
        dict.update(key,self.__TestStepities[key])
    self.cs.put(self._uZED_ATTRIBUTES_ =[("sequential_command","duration",key,expire)]= "//?modified:%r" % key,queue_text = self.__kneeators["expires"])

\end{minted}
\end{minipage}
\\
\\
\hline
\\
1 & 

\begin{minipage}{.9\textwidth}
\begin{minted}[breaklines]{python}
def testPath(self):
    t = Gaffer.Reader(self.callback)
    dupe = ""
    f.mkdir(t)
    f = sys.stdout.tell()
    f.write('_')
    self.assertEqual(f,dataponCollision)

\end{minted}
\end{minipage}
\\
\\
\hline
\\
1 & 

\begin{minipage}{.9\textwidth}
\begin{minted}[breaklines]{python}
def get_count(self):
    return self.get_implicit_count()

\end{minted}
\end{minipage}
\\
\\
\hline
\\
1 & 

\begin{minipage}{.9\textwidth}
\begin{minted}[breaklines]{python}
def is_alive(self):
    return(self.pid,)and(self.pid == 400)

\end{minted}
\end{minipage}
\\
\\
% \\
% \hline
% \\

\bottomrule
\end{tabular}
\caption{\small{Sequences sampled from the original generative model $a$}\label{fab:a_samples}}
\end{table*}

\begin{table*}[t]
\tiny
\begin{tabular}{l|l}
\toprule
\textbf{$b(x)$} & \textbf{Program} \\ 
\midrule
0 & 

% \begin{minipage}{.9\textwidth}
% \begin{minted}[breaklines]{python}
% def ngettext(self,args):
%     return self.diffListView.filter(self.&( OR set))// U>':

% \end{minted}
% \end{minipage}
% \\
% \\
% \hline
% \\
% 1 & 

% \begin{minipage}{.9\textwidth}
% \begin{minted}[breaklines]{python}
% def _callFUT(self):
%     if self.request.MAX_ENCODING == 404:
%         return UserBack
%     return Userarraysize(0)

% \end{minted}
% \end{minipage}
% \\
% \\
% \hline
% \\
% 0 & 

\begin{minipage}{.9\textwidth}
\begin{minted}[breaklines]{python}
def fetch_size(self,page):
    response = self.fetch(page,max((2))
    constant(response.json(),response.pop('utf-8'))
    payload = "%s//%s//%s//%s//%s" %(self.resource.id,page.format_from_bytes())
    return payload

\end{minted}
\end{minipage}
\\
\\
\hline
\\
0 & 

\begin{minipage}{.9\textwidth}
\begin{minted}[breaklines]{python}
def setUp(self):
    self.project_loader = testutil.FileSentenceDependencyGraph(extensions =['file','path'])
    self.schema =RelatedPackage preserveLoader(root_loader)
    self.extension_context = XMLLoader()

\end{minted}
\end{minipage}
\\
\\
\hline
\\
1 & 

\begin{minipage}{.9\textwidth}
\begin{minted}[breaklines]{python}
def __getattr__(self,perm):
    return self._memo.get(perm)

\end{minted}
\end{minipage}
\\
\\
\hline
\\
1 & 

\begin{minipage}{.9\textwidth}
\begin{minted}[breaklines]{python}
def expand(self,text):
    value.strip()
    return extract_cseq(text)

\end{minted}
\end{minipage}
\\
\\
\hline
\\
1 & 

\begin{minipage}{.9\textwidth}
\begin{minted}[breaklines]{python}
def test_Obze(self):
    w = Command()
    self.assertEqual(w.callHeader.callHeader,self.result)

\end{minted}
\end{minipage}
\\
\\
\hline
\\
0 & 

\begin{minipage}{.9\textwidth}
\begin{minted}[breaklines]{python}
def start_stream(self,addressFamily,opcode):
    logger.info("OpenlibwriteStructBegin chunkon.csv',OperationalError())
    error_message = self.get_stream([None,None])
    message,message = self.block_messages[0]
    message = message[0]
    self._process_message(message,message,message,message)

\end{minted}
\end{minipage}
\\
\\
\hline
\\
0 & 

\begin{minipage}{.9\textwidth}
\begin{minted}[breaklines]{python}
def set_dense(self,srs,fit_to):
    if dup in self.scalar:
        return
    if not isinstance(modality,(pyobj):
        self.sq =SUBNET
    self.basic = asim.bin.sample(srs,rng = self.ctypes,trials = self.rng,dtype = self.dtype)

\end{minted}
\end{minipage}
\\
\\
\hline
\\
1 & 

\begin{minipage}{.9\textwidth}
\begin{minted}[breaklines]{python}
def _act(self,value):
    self._result.set_argument('value',value)

\end{minted}
\end{minipage}
\\
\\
\hline
\\
1 & 

\begin{minipage}{.9\textwidth}
\begin{minted}[breaklines]{python}
def _verify_ssling_access_admin(self,ip_name):
    self._check_proxy(ip_name)

\end{minted}
\end{minipage}
\\
\\
\hline
\\
0 & 

\begin{minipage}{.9\textwidth}
\begin{minted}[breaklines]{python}
def __str__(self):
    r =[]
    for s in self.__dict__.items():
        if s[0]in BoundCacheContents():
            break
    if s[:- 1]:Elements([("Unsupported Ct%s]" % ','.join(self.__class__.__name__))
    return "Data attribute '%s' % ','.join("%sCHOICES from %s" %(WARNING,str(r)))

\end{minted}
\end{minipage}
\\
\\
\hline
\\
0 & 

\begin{minipage}{.9\textwidth}
\begin{minted}[breaklines]{python}
def test_FaceIP_3D_14(self):
    self.assertTrue(self.doTestFace(self.doTestFace([self.doTestFace([False,False)])

\end{minted}
\end{minipage}
\\
\\
\hline
\\
0 & 

\begin{minipage}{.9\textwidth}
\begin{minted}[breaklines]{python}
def __init__(self,** options):
    super(_ChoiceTest,self).__init__(** options)
    self.action_classes = options["cells_store"]
    self.choices =(1.2,** options["mysql"]= FakeMissingTuple())
    self.parser = Message(list.__init__(option_forms))

\end{minted}
\end{minipage}
\\
\\
\hline
\\
1 & 

\begin{minipage}{.9\textwidth}
\begin{minted}[breaklines]{python}
def main(self,client):
    remove_home_config(client,"client_snapshot_url")
    self.client.client_snapshot.update(client)

\end{minted}
\end{minipage}
\\
\\
\hline
\\
1 & 

\begin{minipage}{.9\textwidth}
\begin{minted}[breaklines]{python}
def _stop_signal(self,emitter,datafile,for_attachment):
    vim.gui.target_cancel()

\end{minted}
\end{minipage}
\\
\\
% \\
% \hline
% \\

\bottomrule
\end{tabular}
\caption{\small{Sequences sampled from a policy fine-tuned using KL-DPG}\label{fab:dpg_samples}}
\end{table*}

\begin{table*}[t]
\tiny
\begin{tabular}{l|l}
\toprule
\textbf{$b(x)$} & \textbf{Program} \\ 
\midrule
1 & 

% \begin{minipage}{.9\textwidth}
% \begin{minted}[breaklines]{python}
% def __iter__(self):
%     if self._state is None:
%         return self._state
%     else:
%         return None

% \end{minted}
% \end{minipage}
% \\
% \\
% \hline
% \\
% 0 & 

% \begin{minipage}{.9\textwidth}
% \begin{minted}[breaklines]{python}
% def find_fullname(self,dn,sep):
%     try:
%         found =[]
%     except migrations.Dirs:
%         found_assignments =(sxftype.path_p.group(pnt,sep)for rt in found_zones if not found_777215 else created_RUNNING
%     for pure.FileName in found_plugins:
%         cookie_hostname = priv[until_agent]
%         if access_user and 'dst' in self.orig_host.used_users:
%             OlinkroleError('%s:%s' %(download.name,subject),entity_neighbor)

% \end{minted}
% \end{minipage}
% \\
% \\
% \hline
% \\
% 1 & 

\begin{minipage}{.9\textwidth}
\begin{minted}[breaklines]{python}
def invalidateKey(self):
    self.action.rooms = { }

\end{minted}
\end{minipage}
\\
\\
\hline
\\
1 & 

\begin{minipage}{.9\textwidth}
\begin{minted}[breaklines]{python}
def get(self):
    return self.handler.identifier

\end{minted}
\end{minipage}
\\
\\
\hline
\\
1 & 

\begin{minipage}{.9\textwidth}
\begin{minted}[breaklines]{python}
def flush(self):
    self.write("ready")

\end{minted}
\end{minipage}
\\
\\
\hline
\\
1 & 

\begin{minipage}{.9\textwidth}
\begin{minted}[breaklines]{python}
def get_flavor(self,resource,path,** metadata):
    return self.context.get(resource,path,** metadata)

\end{minted}
\end{minipage}
\\
\\
\hline
\\
1 & 

\begin{minipage}{.9\textwidth}
\begin{minted}[breaklines]{python}
def test_api_set_to_result(self):
    X = T.ListHead()
    self.assertEquals(quantiles(X),self._cache.annotations)

\end{minted}
\end{minipage}
\\
\\
\hline
\\
1 & 

\begin{minipage}{.9\textwidth}
\begin{minted}[breaklines]{python}
def is_cmp(self,other):
    return not self._safe_eq(other,self.link)

\end{minted}
\end{minipage}
\\
\\
\hline
\\
1 & 

\begin{minipage}{.9\textwidth}
\begin{minted}[breaklines]{python}
def __iter__(self):
    return iter(self._reverse())

\end{minted}
\end{minipage}
\\
\\
\hline
\\
1 & 

\begin{minipage}{.9\textwidth}
\begin{minted}[breaklines]{python}
def cancel(self):
    return self.enhanced_window.set_timeout()

\end{minted}
\end{minipage}
\\
\\
\hline
\\
1 & 

\begin{minipage}{.9\textwidth}
\begin{minted}[breaklines]{python}
def __str__(self):
    return str(self.repository)

\end{minted}
\end{minipage}
\\
\\
\hline
\\
1 & 

\begin{minipage}{.9\textwidth}
\begin{minted}[breaklines]{python}
def summary(self):
    return self._series

\end{minted}
\end{minipage}
\\
\\
\hline
\\
1 & 

\begin{minipage}{.9\textwidth}
\begin{minted}[breaklines]{python}
def Lazypeer(self):
    return self._peer

\end{minted}
\end{minipage}
\\
\\
\hline
\\
1 & 

\begin{minipage}{.9\textwidth}
\begin{minted}[breaklines]{python}
def ByteSize(self):
    n = 0
    n += self.lengthString(len(self.parameters_))
    return n + self.lengthString(number(self.value_))

\end{minted}
\end{minipage}
\\
\\
\hline
\\
1 & 

\begin{minipage}{.9\textwidth}
\begin{minted}[breaklines]{python}
def setUp(self):
    super(TestMaUserRoleTestCase,self).setUp()
    self.core =BER()
    self.topsetup_existing = False

\end{minted}
\end{minipage}
\\
\\
\hline
\\
1 & 

\begin{minipage}{.9\textwidth}
\begin{minted}[breaklines]{python}
def __init__(self,** kwargs):
    self.sourcemersListComp = kwargs.get('stretch {}'.format(self.__class__.twsourceCentOS_text))

\end{minted}
\end{minipage}
\\
\\
% \hline
% \\

\bottomrule
\end{tabular}
\caption{\small{Sequences sampled from a policy fine-tuned using Reinforce with $R(x) = b(x)$}\label{fab:reinforcer_samples}}
\end{table*}

\begin{table*}[t]
\tiny
\begin{tabular}{l|l}
\toprule
\textbf{$b(x)$} & \textbf{Program} \\ 
\midrule
1 & 

\begin{minipage}{.9\textwidth}
\begin{minted}[breaklines]{python}
def set_OwnerId(self,OwnerId):
    self.add_query_param('OwnerId',OwnerId)

\end{minted}
\end{minipage}
\\
\\
\hline
\\
1 & 

\begin{minipage}{.9\textwidth}
\begin{minted}[breaklines]{python}
def set_OwnerId(self,OwnerId):
    self.add_query_param('OwnerId',OwnerId)

\end{minted}
\end{minipage}
\\
\\
\hline
\\
1 & 

\begin{minipage}{.9\textwidth}
\begin{minted}[breaklines]{python}
def set_OwnerId(self,OwnerId):
    self.add_query_param('OwnerId',OwnerId)

\end{minted}
\end{minipage}
\\
\\
\hline
\\
1 & 

\begin{minipage}{.9\textwidth}
\begin{minted}[breaklines]{python}
def set_OwnerId(self,OwnerId):
    self.add_query_param('OwnerId',OwnerId)

\end{minted}
\end{minipage}
\\
\\
\hline
\\
1 & 

\begin{minipage}{.9\textwidth}
\begin{minted}[breaklines]{python}
def set_OwnerId(self,OwnerId):
    self.add_query_param('OwnerId',OwnerId)

\end{minted}
\end{minipage}
\\
\\
\hline
\\
1 & 

\begin{minipage}{.9\textwidth}
\begin{minted}[breaklines]{python}
def set_OwnerId(self,OwnerId):
    self.add_query_param('OwnerId',OwnerId)

\end{minted}
\end{minipage}
\\
\\
\hline
\\
1 & 

\begin{minipage}{.9\textwidth}
\begin{minted}[breaklines]{python}
def set_OwnerId(self,OwnerId):
    self.add_query_param('OwnerId',OwnerId)

\end{minted}
\end{minipage}
\\
\\
\hline
\\
1 & 

\begin{minipage}{.9\textwidth}
\begin{minted}[breaklines]{python}
def set_OwnerId(self,OwnerId):
    self.add_query_param('OwnerId',OwnerId)

\end{minted}
\end{minipage}
\\
\\
\hline
\\
1 & 

\begin{minipage}{.9\textwidth}
\begin{minted}[breaklines]{python}
def set_OwnerId(self,OwnerId):
    self.add_query_param('OwnerId',OwnerId)

\end{minted}
\end{minipage}
\\
\\
\hline
\\
1 & 

\begin{minipage}{.9\textwidth}
\begin{minted}[breaklines]{python}
def set_OwnerId(self,OwnerId):
    self.add_query_param('OwnerId',OwnerId)

\end{minted}
\end{minipage}
\\
\\
\hline
\\
1 & 

\begin{minipage}{.9\textwidth}
\begin{minted}[breaklines]{python}
def set_OwnerId(self,OwnerId):
    self.add_query_param('OwnerId',OwnerId)

\end{minted}
\end{minipage}
\\
\\
\hline
\\
1 & 

\begin{minipage}{.9\textwidth}
\begin{minted}[breaklines]{python}
def set_OwnerId(self,OwnerId):
    self.add_query_param('OwnerId',OwnerId)

\end{minted}
\end{minipage}
\\
\\
\hline
\\
1 & 

\begin{minipage}{.9\textwidth}
\begin{minted}[breaklines]{python}
def set_OwnerId(self,OwnerId):
    self.add_query_param('OwnerId',OwnerId)

\end{minted}
\end{minipage}
\\
\\
\hline
\\
1 & 

\begin{minipage}{.9\textwidth}
\begin{minted}[breaklines]{python}
def set_OwnerId(self,OwnerId):
    self.add_query_param('OwnerId',OwnerId)

\end{minted}
\end{minipage}
\\
\\
\hline
\\
1 & 

\begin{minipage}{.9\textwidth}
\begin{minted}[breaklines]{python}
def set_OwnerId(self,OwnerId):
    self.add_query_param('OwnerId',OwnerId)

\end{minted}
\end{minipage}
\\
\\
\hline
\\
1 & 

\begin{minipage}{.9\textwidth}
\begin{minted}[breaklines]{python}
def set_OwnerId(self,OwnerId):
    self.add_query_param('OwnerId',OwnerId)

\end{minted}
\end{minipage}
\\
\\
% \hline
% \\

\bottomrule
\end{tabular}
\caption{\small{Sequences sampled from a policy finetuned using Reinforce with $R(x) = P(x)$}\label{fab:reinforceP_samples}}
\end{table*}

\begin{table*}[t]
\tiny
\begin{tabular}{l|l}
\toprule
\textbf{$b(x)$} & \textbf{Program} \\ 
\midrule
    
 \multicolumn{2}{c}{\textbf{Sequences sampled from the original generative model $a$}} \\  
1 & 

\begin{minipage}{.9\textwidth}
\begin{minted}[breaklines]{python}
def closeEvent(self):
    self._isalive = False
    self._original_resume = True

\end{minted}
\end{minipage}
\\
\\
\hline
\\
1 & 

\begin{minipage}{.9\textwidth}
\begin{minted}[breaklines]{python}
def close_file(self):
    pass

\end{minted}
\end{minipage}
\\
\\
\hline
\\
1 & 

\begin{minipage}{.9\textwidth}
\begin{minted}[breaklines]{python}
def closeWorking(self):
    pass

\end{minted}
\end{minipage}
\\
\\
\hline
\\

 \multicolumn{2}{c}{\textbf{Sequences sampled from a policy fine-tuned using KL-DPG}} \\  
1 & 

\begin{minipage}{.9\textwidth}
\begin{minted}[breaklines]{python}
def close(self):
    if not self.closed:
        self.closed = True
    self.translation.close()

\end{minted}
\end{minipage}
\\
\\
\hline
\\
1 & 

\begin{minipage}{.9\textwidth}
\begin{minted}[breaklines]{python}
def close(self):
    self.queue.Importer.close(self.info)
    self.open_input.close()
    self.graph.close(self.gamma)

\end{minted}
\end{minipage}
\\
\\
\hline
\\
1 & 

\begin{minipage}{.9\textwidth}
\begin{minted}[breaklines]{python}
def close(self):
    try:
        self.srv.get_browser.mac(self.bus_process.name,vm_output = True)
    except suspended as ex:
        self.socket.stop(ex)

\end{minted}
\end{minipage}
\\
\\
\hline
\\

 \multicolumn{2}{c}{\textbf{Sequences sampled from a policy fine-tuned using Reinforce with $R(x) = b(x)$}} \\  
1 & 

\begin{minipage}{.9\textwidth}
\begin{minted}[breaklines]{python}
def close(self):
    self._stdout.close()

\end{minted}
\end{minipage}
\\
\\
\hline
\\
1 & 

\begin{minipage}{.9\textwidth}
\begin{minted}[breaklines]{python}
def close(self):
    self.idb.close()

\end{minted}
\end{minipage}
\\
\\
\hline
\\
1 & 

\begin{minipage}{.9\textwidth}
\begin{minted}[breaklines]{python}
def close(self):
    self.reuse = subprocess.Popen('CONNECTION','').unregisterProducer()
    p = subprocess.Popen()
    p.communicate().close()
    return u.close()

\end{minted}
\end{minipage}
\\
\\
\hline
\\

 \multicolumn{2}{c}{\textbf{Sequences sampled from a policy fine-tuned using Reinforce with $R(x) = P(x)$}} \\  
1 & 

\begin{minipage}{.9\textwidth}
\begin{minted}[breaklines]{python}
def close(self,object):
    self.api.close(self.uid.length)

\end{minted}
\end{minipage}
\\
\\
\hline
\\
1 & 

\begin{minipage}{.9\textwidth}
\begin{minted}[breaklines]{python}
def close(self):
    self.job_closed.remove(self)

\end{minted}
\end{minipage}
\\
\\
\hline
\\
1 & 

\begin{minipage}{.9\textwidth}
\begin{minted}[breaklines]{python}
def close(self):
    self.buffer.flush()

\end{minted}
\end{minipage}
\\
\\
% \hline
% \\

\bottomrule
\end{tabular}
\caption{\small{Samples obtained from policies conditioned on prompt \texttt{def close}}\label{tab:def close_samples}}
\end{table*}

\begin{table*}[t]
\tiny
\begin{tabular}{l|l}
\toprule
\textbf{$b(x)$} & \textbf{Program} \\ 
\midrule
    
 \multicolumn{2}{c}{\textbf{Sequences sampled from the original generative model $a$}} \\  
0 & 

\begin{minipage}{.9\textwidth}
\begin{minted}[breaklines]{python}
def fit_pdf(self,hop,theta,theta):
    asserttriangular is self._fit_rewrite(hop, kernel,theta,theta)- gtheta,70)
    assertworkspace isTType.ACCEPTED_ignore
    assert subset in(coeff,Y)
    assert self._Xfd != xOpenStackBackendError
    assert isinstance(750,Win,T,Vector)

\end{minted}
\end{minipage}
\\
\\
\hline
\\
0 & 

\begin{minipage}{.9\textwidth}
\begin{minted}[breaklines]{python}
def fit(self,X,y):
    self._ y = y
    self._children -= 1
    assert isinstance(self._labels,_MOD_'")
    x[:]= 0
    y[:]=Bio_OFFSET
    y *= self._labels
    y * y * y
    y //= y
    return y

\end{minted}
\end{minipage}
\\
\\
\hline
\\
1 & 

\begin{minipage}{.9\textwidth}
\begin{minted}[breaklines]{python}
def fit(self,X = None,y = None,result = None):
    sts = self.get_appId(self.mesh_filename,X,y = y,d = result)
    self.mirror_logpdf([0x9]* indented)

\end{minted}
\end{minipage}
\\
\\
\hline
\\

 \multicolumn{2}{c}{\textbf{Sequences sampled from a policy fine-tuned using KL-DPG}} \\  
1 & 

\begin{minipage}{.9\textwidth}
\begin{minted}[breaklines]{python}
def fit(self,X,y,* args,** kwargs):
    X = self.transform(X,y,* args,** kwargs)
    data = np.DataFrame(data)
    for i in self.fallback_array.iteration_two(* data):
        data[i].labels[i].tolist()
    return data

\end{minted}
\end{minipage}
\\
\\
\hline
\\
0 & 

\begin{minipage}{.9\textwidth}
\begin{minted}[breaklines]{python}
def fit(self, initial_output = None):
    if initial_output:
        self.force_input = False
    else:
        self.cells_done = tuple(initial_output)
    if initial_input == self.WK_MASK:
        self.output_output += self.osfstorage_NORMAL
        self.outputs = list([self.inputState.NORMAL_READ valid])
    return 1

\end{minted}
\end{minipage}
\\
\\
\hline
\\
1 & 

\begin{minipage}{.9\textwidth}
\begin{minted}[breaklines]{python}
def fit(self,reshape,a,b):
    return frappe. filediff(islice(a,b),b)

\end{minted}
\end{minipage}
\\
\\
\hline
\\

 \multicolumn{2}{c}{\textbf{Sequences sampled from a policy fine-tuned using Reinforce with $R(x) = b(x)$}} \\  
1 & 

\begin{minipage}{.9\textwidth}
\begin{minted}[breaklines]{python}
def fit(self,X,y):
    self.x = y

\end{minted}
\end{minipage}
\\
\\
\hline
\\
1 & 

\begin{minipage}{.9\textwidth}
\begin{minted}[breaklines]{python}
def fit(self,fit,d):
    self.fit =followers
    return super(PositionUntilLockedSequence,self).fit(marks)

\end{minted}
\end{minipage}
\\
\\
\hline
\\
0 & 

\begin{minipage}{.9\textwidth}
\begin{minted}[breaklines]{python}
def fit(self,X_acc):
    X_exog = self.xc1.exog
    y = self.instance.exog
    y,= self.model.w2 preserve_uniform(os.environ.XMANllf,y_y))
    y += self.model.t2le continX
    y = self.transition.fit(y)
    y.y = self.model.y * y
    y.red = self.model.gw.urmpopow(y)
    return y

\end{minted}
\end{minipage}
\\
\\
\hline
\\

 \multicolumn{2}{c}{\textbf{Sequences sampled from a policy fine-tuned using Reinforce with $R(x) = P(x)$}} \\  
0 & 

\begin{minipage}{.9\textwidth}
\begin{minted}[breaklines]{python}
def fit(self,fit,X,y,z):
    self.learning = indices[np.zeros(axis = 1Dot,y = y,motion = self. np.loss,y = res.scale)]
    self.index = y

\end{minted}
\end{minipage}
\\
\\
\hline
\\
1 & 

\begin{minipage}{.9\textwidth}
\begin{minted}[breaklines]{python}
def fit(self,params):
    self.params_param = params

\end{minted}
\end{minipage}
\\
\\
\hline
\\
1 & 

\begin{minipage}{.9\textwidth}
\begin{minted}[breaklines]{python}
def fit(self,X,y = None):
    self.x = x
    self.y = x

\end{minted}
\end{minipage}
\\
\\
% \hline
% \\

\bottomrule
\end{tabular}
\caption{\small{Samples obtained from policies conditioned on prompt \texttt{def fit}}\label{tab:def fit_samples}}
\end{table*}

\begin{table*}[t]
\tiny
\begin{tabular}{l|l}
\toprule
\textbf{$b(x)$} & \textbf{Program} \\ 
\midrule
    
 \multicolumn{2}{c}{\textbf{Sequences sampled from the original generative model $a$}} \\  
0 & 

\begin{minipage}{.9\textwidth}
\begin{minted}[breaklines]{python}
def generate_samples_with_prompt(self,input_value,decimal = False):
    use_full = False
    full_input_string = escape_input[decimal]
    newprefix = local_input_format.split("<%s__") % input_label.strip(),[formatted_full])
    return newprefix

\end{minted}
\end{minipage}
\\
\\
\hline
\\
1 & 

\begin{minipage}{.9\textwidth}
\begin{minted}[breaklines]{python}
def generate_samples_with_prompt_publish(self):
    self.overflow = self.percent

\end{minted}
\end{minipage}
\\
\\
\hline
\\
0 & 

\begin{minipage}{.9\textwidth}
\begin{minted}[breaklines]{python}
def generate_samples_with_prompt_line(self):
    lines =[]
    for line in rc:
        if line.startswith('_','-'):
            lines.append("{}0`%s))" % line.replace(".","\n")
            lines.append(": ".join(lines))
    lines.appenddsets()
    lines.append_):
    if len(lines)> 0:
        lines.append(lines[0])
    return lines

\end{minted}
\end{minipage}
\\
\\
\hline
\\

 \multicolumn{2}{c}{\textbf{Sequences sampled from a policy fine-tuned using KL-DPG}} \\  
1 & 

\begin{minipage}{.9\textwidth}
\begin{minted}[breaklines]{python}
def generate_samples_with_prompt(self):
    result = self._generate_blobs().generate(self._name,self._amount_in,lambda x:x.lower())
    return result

\end{minted}
\end{minipage}
\\
\\
\hline
\\
0 & 

\begin{minipage}{.9\textwidth}
\begin{minted}[breaklines]{python}
def generate_samples_with_prompt_token(self,impdly,red,name,restdeclarations,restid_with_mucmapreduce_access_reference,tpversion):
    if prefix_to_acked_ level_per_pbfrom_account_version(MACRO256):
        return 71212000x00 * c201402E64D + 204
    self.generate_cant_rgb_signature(FLAG,name,comtop header, "0|02",["-20001500e6fsha"]

\end{minted}
\end{minipage}
\\
\\
\hline
\\
0 & 

\begin{minipage}{.9\textwidth}
\begin{minted}[breaklines]{python}
def generate_samples_with_prompt(self):
    tsMAIN_SIZE =(0,1)
    tsSBream_bin = self.1000
    if if tsody_size is not None:
        tsbleations = y
    size = ts86.data.get_input(vid_ docs).get_language()
    for address in data.SerializeToString()if not region:
        cpu_ratio = np.zeros(freq.encode("Now"))
        tsLOCATION_examples =[self.read_format(addr)for dir in tsningAssignmentInt()])

\end{minted}
\end{minipage}
\\
\\
\hline
\\

 \multicolumn{2}{c}{\textbf{Sequences sampled from a policy fine-tuned using Reinforce with $R(x) = b(x)$}} \\  
1 & 

\begin{minipage}{.9\textwidth}
\begin{minted}[breaklines]{python}
def generate_samples_with_prompt(self):
    pass

\end{minted}
\end{minipage}
\\
\\
\hline
\\
1 & 

\begin{minipage}{.9\textwidth}
\begin{minted}[breaklines]{python}
def generate_samples_with_prompt_indices(self):
    return self.raw_results_with.raw_options.random_encoding

\end{minted}
\end{minipage}
\\
\\
\hline
\\
0 & 

\begin{minipage}{.9\textwidth}
\begin{minted}[breaklines]{python}
def generate_samples_with_prompt(self,* args,** kwargs):
    return self.fit_sum(kwargs -(n))):

\end{minted}
\end{minipage}
\\
\\
\hline
\\

 \multicolumn{2}{c}{\textbf{Sequences sampled from a policy fine-tuned using Reinforce with $R(x) = P(x)$}} \\  
0 & 

\begin{minipage}{.9\textwidth}
\begin{minted}[breaklines]{python}
def generate_samples_with_prompt(self,cached_done,keep = False):
    if not hasattr(upstream_show,'normalize'):
        return
    sm =wb. cppProcessor(cached_TLS = False)
    self.maxOccurs = self.concurrency. anno_DealList()
    tool.is(csrf_restore,lazy = True)
    self.salt_made(csrf)

\end{minted}
\end{minipage}
\\
\\
\hline
\\
1 & 

\begin{minipage}{.9\textwidth}
\begin{minted}[breaklines]{python}
def generate_samples_with_prompt(self):
    start = back_start - self.start + self.test_samples().set_ofmid
    result =[]
    for step in range(start):
        result.append(step)
        result.append(step)
    return result

\end{minted}
\end{minipage}
\\
\\
\hline
\\
0 & 

\begin{minipage}{.9\textwidth}
\begin{minted}[breaklines]{python}
def generate_samples_with_prompt(self,type::phone_shard = None):
    return int(int(self.last_offsets_best_timescale,type_op = "0"))

\end{minted}
\end{minipage}
\\
\\
% \hline
% \\

\bottomrule
\end{tabular}
\caption{\small{Samples obtained from policies conditioned on prompt \texttt{def generate\_samples\_with\_prompt}}\label{tab:def generate_samples_with_prompt_samples}}
\end{table*}

%\fi % see \iffalse above
\end{document}